\documentclass[accepted]{article}

\usepackage{microtype}
\usepackage{graphicx}
\usepackage{subfigure}
\usepackage{booktabs} 

\usepackage{hyperref}

\usepackage{icml2020}

\usepackage{amsmath, amsfonts, amssymb}
\usepackage{bm}
\usepackage{tikz}
\usetikzlibrary{calc}
\usepackage{pgfplotstable}
\usepackage{pgfplots}
\usepackage{array,multirow,graphicx}
\usepackage{tabularx,colortbl}
\usepackage[colorinlistoftodos,prependcaption]{todonotes}
\usepackage{enumitem}
\usepackage{standalone}
\pgfplotsset{compat=1.13}

\hypersetup{
    pdfproducer={},  % producer of the document
}

\renewcommand{\vec}[1]{\boldsymbol{#1}}
\newcommand{\mat}[1]{\boldsymbol{\mathrm{#1}}}

\def \y{\mathbf{y}}

\def \R{\mathbb{R}}			
\def \N {\mathbb{N}}

\def \I{\mat{I}}			

\DeclareMathAlphabet{\pazocal}{OMS}{zplm}{m}{n}

\newcommand{\algorithmicconstants}{\textbf{Constants:}}
\newcommand\CONSTANTS{\item[\algorithmicconstants]}
\newcommand{\algorithmicreturn}{\textbf{return: }}
\newcommand\RETURN{\item[\algorithmicreturn]}

\newcommand{\algorithmiccommentMine}[1]{\bgroup\hfill$\triangleright$~\emph{#1}\egroup}

\usepackage{amsthm}

\usetikzlibrary{calc}
\usepackage{stmaryrd}
\usepackage{upgreek}
\usetikzlibrary{shapes.geometric}
\usetikzlibrary{decorations.pathreplacing}

\definecolor{myblack}{RGB}{53, 53, 53}
\definecolor{myblue}{RGB}{40, 75, 99}
\definecolor{myred}{RGB}{192, 50, 33}
\definecolor{myyellow}{RGB}{255, 166, 48}
\definecolor{mywhite}{RGB}{240, 237, 238}
\definecolor{mygreen}{RGB}{0, 102, 0}

\definecolor{green1}{RGB}{9, 82, 86}
\definecolor{green2}{RGB}{8, 127, 140}
\definecolor{green3}{RGB}{6, 167, 125}
\definecolor{green4}{RGB}{79, 109, 122}
\definecolor{green5}{RGB}{192, 214, 223}
\definecolor{violet}{RGB}{26,69,131}

\newcommand\oldtext[1]{}

\icmltitlerunning{Multilevel Minimization for Deep Residual Networks}

\begin{document}

\twocolumn[
\icmltitle{Multilevel Minimization for Deep Residual Networks}

\icmlsetsymbol{equal}{*}

\begin{icmlauthorlist}
\icmlauthor{Lisa Gaedke-Merzh\"auser}{equal,usi}
\icmlauthor{Alena Kopani\v{c}\'akov\'a}{equal,usi}
\icmlauthor{Rolf Krause}{usi}
\end{icmlauthorlist}

\icmlaffiliation{usi}{Institute of Computational Science, Universit\`a della Svizzera italiana}

\icmlcorrespondingauthor{Alena Kopani\v{c}\'akov\'a}{alena.kopanicakova@usi.ch}

\icmlkeywords{deep residual networks, optimal control problem, multilevel minimization}

\vskip 0.3in
]

\printAffiliationsAndNotice{\icmlEqualContribution} 

\begin{abstract}
We present a new multilevel minimization framework for the training of deep residual networks (ResNets), which has the potential to significantly reduce  training time and effort. 
Our  framework is based on the dynamical system's viewpoint, which formulates a ResNet as the discretization of an initial value problem.
The training process is then formulated as a time-dependent optimal control problem, which we discretize using different time-discretization parameters, 
eventually generating  multilevel-hierarchy of auxiliary networks with different resolutions.
The training of the original ResNet is then enhanced by training the auxiliary networks with reduced resolutions. 
By design, our framework is conveniently independent of the choice of the training strategy chosen on each level of the multilevel hierarchy. 
By means of numerical examples, we analyze the convergence behavior of the proposed method and demonstrate its robustness.
For our examples we employ a multilevel gradient-based methods. 
Comparisons with standard single level methods show a speedup of more than factor three while achieving the same validation accuracy.
\end{abstract}

\section{Introduction}
\label{sec:introduction}
Deep residual networks or ResNets are widely used architectures that demonstrate state-of-the-art performance in complex statistical learning tasks with applications in various fields, such as computer vision \cite{jung2017resnet, chen2017deeplab}, or speech recognition \cite{wu2016google, xiong2018microsoft}. 
The popularity of ResNets originates from their remarkable performance in the ImageNet \cite{ILSVRC15} and the MS COCO \cite{lin2014microsoft} image recognition competitions. 

A major drawback of very deep ResNets is their long training time.
To mitigate this issue, different strategies have been proposed, for example
networks with stochastic depth \cite{huang2016deep}, 
mollifying networks \cite{gulcehre2016mollifying},  
spatially adaptive architectures \cite{figurnov2017spatially}, 
or multilevel parameter initialization strategies \cite{haber2018learning, chang2017multi}.

In this work, we propose to accelerate the training of ResNets using multilevel minimization.
Our work is motivated by the fact that the network depth is of paramount importance for achieving the necessary approximation properties \cite{haastad1991power, simonyan2014very}.
However, very deep networks are computationally expensive to train, as the cost of forward-backward propagation scales linearly with respect to the number of parameters. 
In contrast, shallower networks might not show the necessary approximation properties, but their training cost is relatively low.
Our multilevel framework exploits a multilevel hierarchy of auxiliary networks with different depths.
The training of the deepest network is then accelerated by internally training the shallower networks.

The proposed multilevel framework is inspired by multigrid methods \cite{Briggs2000multigrid, hackbusch2013multi}, which have originally been developed for the solution of elliptic partial differential equations. 
An extension of linear multigrid methods to nonlinear problems, called full approximation scheme (FAS), can be found in \cite{brandt1977multi}.
Later, several nonlinear multilevel minimization techniques have emerged, for example the multilevel line-search method (MG/OPT) \cite{Nash2000multigrid}, the recursive multilevel trust region method  (RMTR) \cite{Gratton2008recursive, Gross2009, kopanicakova2020b, kopanicakova2020a}, or
higher-order multilevel optimization strategies \cite{calandra2019high}.
Our multilevel minimization method can be seen as a variant of an MG/OPT framework which is tailored for training ResNets.

The main challenge in designing an efficient multilevel minimization framework is to construct a suitable multilevel hierarchy.
Here, we leverage the emerging dynamical system's viewpoint  \cite{haber2018learning, weinan2017proposal}, which casts a ResNet as the discretization of an initial value problem.
The training process is then formulated as the minimization of a time-dependent optimal control problem.
As a consequence, we can obtain a hierarchy of ResNets with different depths by discretizing the same optimal control problem with different discretization parameters.

A dynamical system's viewpoint was first used in a multilevel context in  \cite{chang2017multi},  where the authors trained shallow networks to initialize parameters of a deep network.
The same parameter initialization strategy was recently extended for layer-parallel training of ResNets \cite{cyr2019multilevel}.
Our method differs from the methods proposed in \cite{chang2017multi} and \cite{cyr2019multilevel}, as we take advantage of a multilevel hierarchy during the whole training process, not only in the beginning.
Nevertheless, it is possible to incorporate a multilevel initialization strategy into our multilevel minimization framework.
We do not exploit this possibility in the presented paper, as aim of this work is to test the proposed multilevel training framework by itself.

This work makes the following contributions:
\begin{itemize}[noitemsep, topsep=0pt]
\item We present an abstract nonlinear multilevel minimization framework for training deep residual networks.  
\item Using our multilevel framework, we propose multilevel variants of gradient and mini-batch gradient methods.
\item We numerically analyze the convergence behavior of our multilevel training strategies using two different datasets and ResNets with more than $2,000$ layers. 
In addition, comparisons with a standard single level methods are made, which demonstrate a speed-up of more than factor three.
\end{itemize}

 \section{Deep Residual Networks}
\label{sec:resnetdynamical}
This section provides a brief overview of deep residual networks (ResNets) in the context of supervised classification.  
Through the following, we consider a dataset $\mathcal{D} = \{(\vec{x}_j, \vec{c}_j)\}_{j=1}^p$ of $p$ samples. 
Each sample is a pair consisting of an input feature $\vec{x}_j \in \R^q$ and its corresponding label $\vec{c}_j \in \R^{m}$.
The size of the label vector $\vec{c}_j$ is determined by the number of output classes, i.e. $m$, as the $i$-th component of vector $\vec{c}_j$ corresponds to the probability of example $\vec{x}_j$ belonging to the $i$-th class.

\subsection{Classification}
The main idea behind supervised learning is to construct a model, which describes the relationship between input and output for a labeled dataset $\mathcal{D}$. 
The model function $f_m: \R^q \times \R^n \rightarrow \R^m$ is parametrized by a set of parameters $\vec{\uptheta} \in \R^n$.
The process of finding suitable parameters $\vec{\uptheta}$ is called training and it usually requires solving the following minimization problem:
 \begin{equation}
\begin{aligned}
& \min_{\vec{\uptheta}}  \frac{1}{p} \sum_{j=1}^{p} \ell(f_m(\vec{x}_j, \vec{\uptheta}), \vec{c}_j) +\mathcal{R}(\vec{\uptheta}),
\end{aligned}
\label{eq:generic_min_problem}
\end{equation}
where a loss function $\ell: \R^m \rightarrow \R$ measures the deviation of the the predicted output from the known label.
The regularizer $\mathcal{R}: \R^n \rightarrow \R$ in \eqref{eq:generic_min_problem} is chosen such that it ensures the existence and regularity of the parameters $\vec{\uptheta}$. 
A common choice for the regularizer is Tikhonov regularization \cite{engl1996regularization}, however other possibilities have also been used, see for example \cite{ng2004feature}.
\oldtext{The training is considered effective if the obtained parameters $\vec{\uptheta}$ generalize well to previously unseen data. }

In the context of classification, the model function $f_m$ is constructed by composing the forward propagation $f: \R^q \times \R^r \rightarrow \R^v$ with the hypothesis function $\mathcal{P}: \R^{m} \rightarrow \R^{m}$.
The forward propagation filters input features in a nonlinear manner, while the hypothesis function predicts the class label probabilities using the output of the forward propagation.
In abstract form, the model function $f_m$ is defined as
\begin{align}
f_m(\vec{x}, \vec{\uptheta}) = \mathcal{P}(\mat{W}_K f(\vec{x}, \vec{\theta}) + \vec{b}_K),
\label{eq:f_composition}
\end{align}
where we split the model parameters $\vec{\uptheta}$ into parameters of classification $\vec{\theta}_K$ and forward propagation $\vec{\theta}$, thus $\vec{\uptheta}= \{\vec{\theta}, \vec{\theta}_K \}$.
The classification parameters $\vec{\theta}_K=\{ \mat{W}_K, \vec{b}_K\}$ consist of weights $\mat{W}_K \in \R^{m \times v}$ and biases $\vec{b}_K \in \R^{m}$.
For multinomial classification problems, \oldtext{i.e. problems where the number of output classes is bigger than two,} it is common to employ a cross-entropy loss function
together with the softmax hypothesis function.
For alternatives choices, we refer interested readers to \cite{goodfellow2016deep}. 

\subsubsection{Forward propagation via ResNet}
In deep learning, the neural network constitutes a form of forward propagation function $f$. 
The parametric function $f$ is created by concatenating many functions, called layers. 
Each layer $k$ is usually composed of affine linear and point-wise nonlinear transformations, that are parametrized by the layer parameters $\vec{\theta}_k \in \R^d$.

In this work, we consider residual networks with identity shortcut connections \cite{he2016identity}. 
The propagation of the input sample $\vec{x}$ through a network  with $K$ residual layers can be then expressed as
\begin{equation}
\vec{y}_{k+1} = \vec{y}_{k} + \mathcal{F}(\vec{y}_{k}, \bm{\theta}_{k}),  \ k \in \{0, \dots, K-1\},
\label{eq:res_layer}
\end{equation}
where $\vec{y}_{k} \in \R^{v}$ denotes the state of layer $k$. 
For simplicity, Equation \eqref{eq:res_layer} assumes a constant network width $v$.
Hence, we map an input sample $\vec{x} \in \R^{q}$ into the feature space with the help of the linear operator $\mat{Q} \in \R^{v \times q}$, e.g. $\mat{y}_0 := \mat{Q} \vec{x}$.
The elements of the matrix $\mat{Q}$ can be fixed or learned during the training process. 

The transformation $\mathcal{F}: \R^{v} \times \R^d \rightarrow \R^v$ from \eqref{eq:res_layer}  describes the residual module, c.f. \cite{he2016deep}. 
Here, we assume that $\mathcal{F}$ takes form of the simple one layer perceptron
\begin{align}
\mathcal{F}(\vec{y}_{k}, \vec{\theta}_{k}) := \sigma(\mat{W}_{k} \vec{y}_{k}  + \vec{b}_{k}),
\label{eq:f_tilde}
\end{align}
where $\sigma: \R^{v} \rightarrow \R^{v}$ is the nonlinear activation function, for example the rectified linear unit (ReLu), defined as $\sigma(\vec{z}) := \max\{\vec{0}, \mathbf{z}\}$. For alternatives, such as logistic sigmoid, or hyperbolic tangent, see \cite{goodfellow2016deep}. 
The affine transformations in \eqref{eq:f_tilde} are defined by a set of layer parameters $\vec{\theta}_k:= \{ \mat{W}_k, \vec{b}_k \}$ consisting of weights $\mat{W}_k \in \R^{v \times v}$ and biases $\vec{b}_k \in \R^v$.
The linear operator $\mat{W}_k$ can be a dense matrix, or sparse, e.g. in the case of a convolutional neural network, where it expresses the convolutional operator, see \cite{goodfellow2016deep}.

\subsection{Classification as optimal control problem}
Following  \cite{haber2018learning}, Equation \eqref{eq:res_layer} can be seen as a simplification of the more generic formula for a one-step method
\begin{align}
\vec{y}_{k+1} = \vec{y}_{k} + \Delta_t   \mathcal{F}(\vec{y}_{k}, \vec{\theta}_{k}), 
\label{eq:f_tilde_h}
\end{align}
with $\Delta_t = 1$.
Now, the forward propagation through the network \eqref{eq:f_tilde_h} can be interpreted as a forward Euler discretization of the initial value problem
\begin{equation}
\begin{aligned}
\partial_t \vec{y}(t) &= \mathcal{F}(\vec{y}(t), \vec{\theta}(t)),  & \quad  \forall t \in (0,T], \\
\vec{y}(0) &= \mat{Q} \vec{x}.  & 
\end{aligned}	
\label{eq:dynamical_system}
\end{equation}
The dynamical system above then continuously transforms the initial state $\vec{y}(0)$ into the network output $\vec{y}(T)$, while 
the time-dependent control variables $\vec{\theta}(t)$ define the behavior of the system. 
The classification problem is now formulated as the following continuous optimal control problem \cite{haber2017stable}:
\begin{align}
 &\min_{\vec{\theta}, \vec{\theta}_K, \vec{y}} \frac{1}{p} \sum_{j=1}^{p}\ell(\mathcal{P}(\mat{W}_K \vec{y}_j(T) + \vec{b}_K), \vec{c}_j) + \int\limits_{0}^{T} \mathcal{R}(\vec{\theta}(t), \vec{\theta}_K )\nonumber \\  
&\quad   \text{subject to}  \quad \partial_t \vec{y}_j(t) = \mathcal{F}(\vec{y}_j(t), \vec{\theta}(t)),  \label{eq:cts_problem}  \\
& \quad  \quad \quad \quad \quad \ \ \quad \vec{y}_j(0) = \mat{Q}\vec{x}_j,\nonumber
\end{align}
where $\vec{y}_j(T)$ denotes the output of the network for the data sample $\vec{x}_j$. 
The continuous formulation \eqref{eq:cts_problem} opens the door to many new developments. 
For example, the design of stable network architectures \cite{haber2017stable, benning2019deep}, the parallel approach to training \cite{gunther2018layer, parpas2019predict}, or novel solution strategies \cite{li2017maximum}.
In this work, we leverage the continuous formulation in order to design an efficient multilevel training strategy, see Section \ref{sec:multigrid}. 

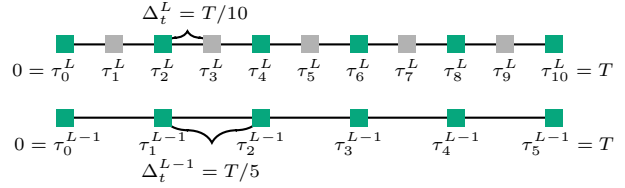
\begin{figure}
	\centering
    \begin{tikzpicture}[scale=0.65, font=\scriptsize]
	\draw[-, thick] (0,0)  -- (10, 0);

     \draw [thick, decorate,decoration={brace,amplitude=5pt, raise=0pt},xshift=0.0cm,yshift=0pt]
(2, 0)  -- (3, 0) node [black,xshift=-0.225cm, yshift=12.0]{ $\Delta_t^L = T/10$};	

       \foreach \x in  {0,2,4, 6, 8,10} 
          \node at (\x, 0) [rectangle,fill=green3]  {};      
          
       \foreach \x in {1, 3, 5, 7, 9} 
          \node at (\x, 0) [rectangle,fill=gray!60!]  {};

       \foreach \x in  {1,...,9} 
         \node [below] at (\x, -0.1) {$\tau_{\x}^L$};

     \node [below] at (0,-0.1) {$0=\tau_{0}^L$ \enspace \enspace \enspace \enspace };     
     \node [below] at (10,-0.1) {\enspace \enspace \enspace \enspace \enspace $\tau_{10}^L=T$};

\draw [thick, decorate,decoration={brace,amplitude=9.5pt, mirror, raise=0pt},xshift=0.0cm,yshift=0pt]
(2, -1.5)  -- (4, -1.5) node [black,xshift=-0.8cm, yshift=-20.0]{$\Delta_t^{L-1} = T/5$};	

	\draw[-, thick] (0,-1.5)  -- (10, -1.5);
         
       \foreach \x in  {0,2,4, 6, 8,10} 
          \node at (\x, -1.5) [rectangle,fill=green3]  {};      
         
         \node [below] at (2, -1.6) {$\tau_{1}^{L-1}$};
         \node [below] at (4, -1.6) {$\tau_{2}^{L-1}$};
         \node [below] at (6, -1.6) {$\tau_{3}^{L-1}$};
         \node [below] at (8, -1.6) {$\tau_{4}^{L-1}$};                           
         
     \node [below] at (0, -1.6) {$0=\tau_0^{L-1}$  \enspace };         
     \node [below] at (10,-1.6) {\enspace \enspace \enspace $\tau_{5}^{L-1}=T$};                           
	
    \end{tikzpicture}
    \caption{An example of time grids used for the multilevel discretization.
    On the fine level, we consider $10$ time-steps, while on the coarse level, we use $5$ larger time-steps.}
    \label{fig:time_intervals}
\end{figure}

\subsubsection{Discretization}
\label{sec:discretization}
To solve the continuous optimal control problem \eqref{eq:cts_problem} numerically, we discretize \eqref{eq:cts_problem} in time. 
Thus, we consider the time-grid $0 = \tau_0 <  ....< \tau_K = T$ of $K+1$ uniformly distributed time points $\tau_k:= \Delta_t k$, where $\Delta_t: = T/K$ represents a time-step. 
The discretized control $\vec{\theta}_k \approx \vec{\theta}(\tau_k)$ and state $\vec{y}_{j,k} \approx \vec{y}_j(\tau_k)$ variables then correspond to the parameters and the state of the $k$-th layer of the ResNet, respectively. 

In the discrete setting, we obtain the following constrained minimization problem:
\begin{align}
& \min_{\vec{\theta}, \vec{\theta}_K,  \vec{y}} \underbrace{  \frac{1}{p}  \sum_{j=1}^{p} \ell(\mathcal{P}(\mat{W}_K \vec{y}_{j,K} + \vec{b}_K), \vec{c}_j)   + \sum_{k=0}^{K-1} \mathcal{R}(\vec{\theta}_k, \vec{\theta}_K)}_{=: \ \mathcal{L}(\vec{\theta}, \vec{\theta}_K, \vec{y}_K) \ := \mathcal{L}(\vec{\uptheta}, \vec{y}_K)} \nonumber \\
 &\text{subject to }  \quad \vec{y}_{j,k+1} = \vec{y}_{j,k} + \Delta_t F(\vec{y}_{j,k}, \vec{\theta}_k), \label{eq:discrete_problem}  \\
&  \hspace{1.85cm} \quad \mathbf{y}_{j,0} = \mat{Q} \vec{x}_j, \nonumber
\end{align}
where we have used an explicit Euler scheme to discretize the time derivative $\partial_t \vec{y}(t)$ in  \eqref{eq:cts_problem}.
This choice of discretization is what imposes the particular ResNet architecture.
However, other, possibly more stable discretization schemes, can be considered, see for instance \cite{haber2017stable}. 
Employing an explicit Euler method, we can ensure the stability of a forward propagation by ensuring that the time-step $\Delta_t$ is sufficiently small \cite{haber2017stable}.

\paragraph{Multilevel discretization}
We can discretize \eqref{eq:cts_problem} using different discretization parameters.
This allows us to construct a multilevel-hierarchy of auxiliary networks with different resolutions.
We consider a hierarchy of $L$ levels, denoted by $l = 1, \dots, L$. 
The finest level, $l=L$, represents the discretization of the optimal control problem \eqref{eq:cts_problem} with satisfactory resolution/representation capacity.

This means, that the time-step $\Delta_t^L$ is sufficiently small and that the network has sufficiently many layers to ensure desirable approximation properties of the model. 
In order to obtain coarser level networks, we discretize the time interval $[0, T]$ with larger time-steps.
For instance, if we assume a uniform coarsening in time by a factor of two, the following relation holds for time-steps of subsequent levels $\Delta_t^{l-1}=2\Delta_t^{l}$. 
As a consequence, the number of layers is halved between the networks on level $l$ and $l-1$.
Figure~\ref{fig:time_intervals} demonstrates the process for a simple $2$-level example. 
Since the networks on coarser levels of the multilevel hierarchy are constructed with fewer layers, they have less trainable parameters.
Therefore, they are computationally cheaper to optimize, due to the fact that the cost of forward-backward propagation used during the training grows linearly with respect to the number of parameters \cite{hecht1992theory}. 
As a consequence, it is roughly two--times faster to perform one forward--backward propagation on a coarser level than on the subsequent finer level.

 \section{Multilevel Training for ResNets}
\label{sec:multigrid}
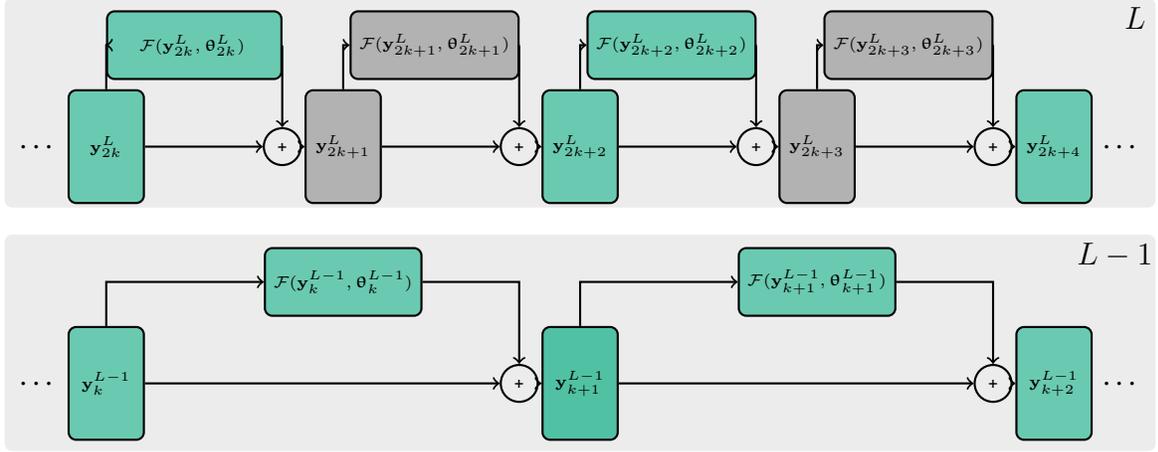
\begin{figure*}
	\centering
    \begin{tikzpicture}[scale=0.9, square/.style={regular polygon,regular polygon sides=4, minimum width=1.5cm}, circlenode/.style={circle, draw, minimum size=0.01cm}, squareRB/.style={rounded corners=0.1cm, minimum width=2cm,minimum height=0.9cm}, squareY/.style={rounded corners=0.1cm, minimum width=1cm,minimum height=1.5cm}, font=\tiny]

	\fill [gray!15, rounded corners=0.1cm] (-1.5,3.1) rectangle (15.5, 6.2);	
	\fill [gray!15,  rounded corners=0.1cm] (-1.5, -0.5) rectangle (15.5, 2.7);		
	
	\node at (15.2, 5.9){\large $L$};
	\node at (14.9, 2.4){\large $L-1$};			

	\node  [squareY,draw, thick, fill=green3!60!] (1) at (0,4) { $\y_{2k}^L$};
	\node  [squareRB,draw, thick, fill=green3!60!] (2) at (1.3,5.5) {\enspace  $ \enspace \enspace \mathcal{F}(\y_{2k}^L, \vec{\uptheta}_{2k}^L)$ \enspace \enspace \enspace };	
	
	\node  [squareY,draw, thick,  fill=gray!60!] (3) at (3.5,4) {$\y_{2k+1}^L$};
	\node  [squareRB,draw, thick, fill=gray!60!] (4) at (4.85,5.5) { $\mathcal{F}(\y_{2k+1}^L, \vec{\uptheta}_{2k+1}^L)$};		
		
	\node  [squareY,draw, thick, fill=green3!60!] (5) at (7,4) {$\y_{2k+2}^L$};
	\node  [squareRB,draw, thick, fill=green3!60!] (6) at (8.35,5.5) { $\mathcal{F}(\y_{2k+2}^L, \vec{\uptheta}_{2k+2}^L)$};		
		
	\node  [squareY,draw, thick, fill=gray!60!] (7) at (10.5,4) {$\y_{2k+3}^L$};
	\node  [squareRB,draw, thick, fill=gray!60!] (8) at (11.85,5.5) { $\mathcal{F}(\y_{2k+3}^L, \vec{\uptheta}_{2k+3}^L)$};		
		 
	\node  [squareY,draw, thick, fill=green3!60!] (9) at (14,4) {$\y_{2k+4}^L$};

	\node  [circlenode,draw, thick] (c1) at (2.6,4) {\textbf{+}};	
	\node  [circlenode,draw, thick] (c2) at (6.1,4) {\textbf{+}};	
	\node  [circlenode,draw, thick] (c3) at (9.6,4) {\textbf{+}};	
	\node  [circlenode,draw, thick] (c4) at (13.1,4) {\textbf{+}};	

	\node at (-1.0, 4){\large \dots};
	\node at (15, 4){\large \dots};

	\draw[->, thick](1)  -- (0, 5.5) -- (2);	
	\draw[->, thick](3)  -- (3.5, 5.5) -- (4);	
	\draw[->, thick](5)  -- (7, 5.5) -- (6);	
	\draw[->, thick](7)  -- (10.5, 5.5) -- (8);

	\draw[->, thick](2)  -- (2.6, 5.5) -- (c1);	
	\draw[->, thick](4)  -- (6.1, 5.5) -- (c2);	
	\draw[->, thick](6)  -- (9.6, 5.5) -- (c3);	
	\draw[->, thick](8)  -- (13.1, 5.5) -- (c4);

	\draw[->, thick] (1) -- (c1);
	\draw[->, thick] (3) -- (c2);
	\draw[->, thick] (5) -- (c3);
	\draw[->, thick] (7) -- (c4);					

	\draw[->, thick] (c1) -- (3);
	\draw[->, thick] (c2) -- (5);
	\draw[->, thick] (c3) -- (7);
	\draw[->, thick] (c4) -- (9);

	\node  [squareY,draw, thick, fill=green3!60!] (1c) at (0,0.5) { $\y_{k}^{L-1}$};
	\node  [squareRB,draw, thick, fill=green3!60!] (2c) at (3.5,2) { $\mathcal{F}(\y_{k}^{L-1}, \vec{\uptheta}_{k}^{L-1})$};	
		
	\node  [squareY,draw, thick, fill=green3!70!] (5c) at (7,0.5) {$\y_{k+1}^{L-1}$};
	\node  [squareRB,draw, thick, fill=green3!60!] (6c) at (10.5,2) { $\mathcal{F}(\y_{k+1}^{L-1}, \vec{\uptheta}_{k+1}^{L-1})$};		
		 
	\node  [squareY,draw, thick, fill=green3!60!] (9c) at (14,0.5) {$\y_{k+2}^{L-1}$};
	\node  [circlenode,draw, thick] (c2c) at (6.1,0.5) {\textbf{+}};	
	\node  [circlenode,draw, thick] (c4c) at (13.1,0.5) {\textbf{+}};	

	\node at (-1.0, 0.5){\large \dots};
	\node at (15, 0.5){\large \dots};		

	\draw[->, thick](1c)  -- (0, 2) -- (2c);	
	\draw[->, thick](5c)  -- (7, 2) -- (6c);	

	\draw[->, thick](2c)  -- (6.1,2) -- (c2c);	
	\draw[->, thick](6c)  -- (13.1, 2) -- (c4c);							

	\draw[->, thick] (1c) -- (c2c);
	\draw[->, thick] (5c) -- (c4c);					

	\draw[->, thick] (c2c) -- (5c);
	\draw[->, thick] (c4c) -- (9c);

    \end{tikzpicture}
    \caption{An example of a multilevel hierarchy of ResNets. The state and control variables are discretized using different time grids.}
    \label{fig:ml_resnets}
\end{figure*}

In this section, we introduce a nonlinear multilevel minimization framework for training ResNets. 
The presented framework can be seen as a variant of the MG/OPT framework \cite{Nash2000multigrid} originally developed for solving the large scale problems arising from the discretization of partial differential equations. 
Our variant of MG/OPT is tailored to the minimization of the discrete optimal control problem \eqref{eq:discrete_problem}.
In particular, we employ a hierarchy of auxiliary networks with different depths, see Figure \ref{fig:ml_resnets}, which are used to accelerate the training of the original network. 
Each auxiliary network is trained by approximately minimizing the associated level-dependent optimal control problem, see Section \ref{sec:level_dep_op} for the details. 
The minimization of the level-dependent optimal control problem is carried out using an optimizer associated with a given level.

Through the following, we use a pair $(l, \mu^l)$ of superscripts to denote the quantities related to a level $l$ and iteration $\mu^l$.
If no subscript is used, we refer to quantities on all layers of the network simultaneously. 
Otherwise, the subscript identifies the quantities associated with a given layer.
For example, $\vec{\uptheta}_k^{1,\mu^1}$ denotes the parameters related to the $k$-th layer of the coarsest network, $l=1$, after $\mu^1$ update steps.

\paragraph{Transfer operators}
The multilevel training framework requires to transfer data between subsequent levels of the multilevel hierarchy. 
For this reason, we employ two types of transfer operators. 
The interpolation operator $\mat{I}^{l}: \R^{n^l} \rightarrow \R^{n^{l+1}}$ transfers weights and biases from level $l$ to level $l+1$. 
Here, we consider piecewise constant interpolation in time.
Other choices of the transfer operators, such as linear interpolation, are also possible and may be even preferable.
We plan to incorporate them into our multilevel training framework in future work. 
In addition to the interpolation operator, the multilevel method also uses a restriction operator $\mat{R}^l: \R^{n^{l+1}} \rightarrow \R^{n^{l}}$, in order to transmit data, 
such as gradients, from level $l+1$ to level $l$. 
As common in multgrid literature \cite{hackbusch2013multi}, we choose the restriction operator as $\mat{R}^l := (\mat{I}^{l})^T$.

\subsection{Multilevel training}
\label{sec:ml_training}
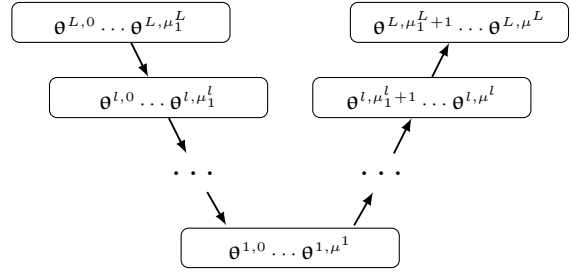
\begin{figure}[H]
\centering
\begin{tikzpicture}[samples=500, font=\scriptsize, squareRB/.style={rounded corners=0.1cm, minimum width=2.9cm,minimum height=0.5cm}, emptyBox/.style={rounded corners=0.1cm, minimum width=0.01cm,minimum height=0.5cm}]
	
	\node[squareRB, draw] (v2) at (1, 2) {$\vec{\uptheta}^{L,0} \cdots \vec{\uptheta}^{L,\mu^L_1}$};
	\node[squareRB, draw] (v22) at (5.5, 2) {$\vec{\uptheta}^{L,\mu_1^L+1} \cdots \vec{\uptheta}^{L,\mu^{L}}$};
		
	\node[squareRB, draw] (v1) at (1.5, 1) {$\vec{\uptheta}^{l,0} \cdots \vec{\uptheta}^{l,\mu_1^l}$};
	\node[squareRB, draw] (v11) at (5, 1) {$\vec{\uptheta}^{l,\mu_1^l +1} \cdots \vec{\uptheta}^{l, \mu^{l}}$};	

	\node[squareRB] (empty) at (2, 0) { \Large $\dots$};
	\node[squareRB] (empty2) at (4.5, 0) {\Large $\dots$};	
	
	\node[squareRB, draw] (v0) at (3.25, -1) {$\vec{\uptheta}^{1,0} \cdots \vec{\uptheta}^{1,\mu^1}$};

	\draw[color=black,line width=0.25mm, -latex] (v2)--(v1); 	
	\draw[color=black,line width=0.25mm, -latex] (v1)--(empty); 		
	\draw[color=black,line width=0.25mm, -latex] (empty)--(2.4, -0.7);

	\draw[color=black,line width=0.25mm, -latex] (4.1, -0.7)--(empty2); 	
	\draw[color=black,line width=0.25mm, -latex] (empty2)--(v11); 		
	\draw[color=black,line width=0.25mm, -latex] (v11)--(v22); 	
	
	\end{tikzpicture}
	\caption{A scheme of the V-cycle used during multilevel training.}
	\label{fig:V_cycle}
\end{figure}

The MG/OPT iteration has the form of a V--cycle, which consists of a downward and an upward phase, see Figure \ref{fig:V_cycle}.
The downward phase starts on the finest level, $l=L$, with initial weights $\vec{\uptheta}^{L,0}$ and passes through all levels until the coarsest level is reached. 
On each level, we perform $\mu^l_1$ level-optimizer steps in order to find an approximate solution of the level-dependent optimal control problem.  
The approximate solution, i.e. the updated network parameters $\vec{\uptheta}^{l,\mu_1^l}$ are then used to initialize weights on the subsequent coarser level, e.g. $\vec{\uptheta}^{l-1,0}=\mat{R}^{l-1} \vec{\uptheta}^{l,\mu_1^l}$. 
This process is repeated until we reach the coarsest level, $l=1$.

Once the coarsest level is reached and we have performed $\mu^1$ level-1-optimizer step, yielding the parameters $\vec{\uptheta}^{1, \mu^1}$, we can initiate the upward phase.
During the upward phase, we return to the finest level, while passing through all levels of the multilevel hierarchy. 
Starting on the coarsest level, we compute the coarse grid correction $\vec{e}^{l} = \vec{\uptheta}^{l,\mu^l} - \vec{\uptheta}^{l,0}$, which characterizes the difference between the initial and the updated parameters on a given level. 
This correction is then transferred to the next finer level using the interpolation operator as $\vec{e}^{l+1} = \mat{I}^l \vec{e}^{l}$. 
Once we have the interpolated correction, we use it to update the parameters of the finer network, thus $\vec{\uptheta}^{l+1,\mu_1^{l+1} +1} = \vec{\uptheta}^{l+1,\mu_1^l} + \vec{e}^{l+1}$.
Finally, we perform $\mu_2^l$ steps of the level-optimizer in order to improve the current approximation of the parameters on level $l$.
The whole process is summarized in Algorithm \ref{alg:mgopt}. 

\subsubsection{Level-dependent minimization problems}
\label{sec:level_dep_op}
On each level of the multilevel hierarchy, we look for an approximate solution of some level-dependent optimal control problem. 
As common for nonlinear multilevel (minimization) schemes, such as FAS \cite{brandt1977multi}, or RMTR \cite{Gratton2008recursive}, we define the level-dependent optimal control problems as
\begin{align}
 \min_{\vec{\uptheta}^l, \vec{y}^l}  \quad  \mathcal{H}^l(\vec{\uptheta}^l, \vec{y}^l_K) :&=    \mathcal{L}^l(\vec{\uptheta}^l, \vec{y}^l_K) +  \langle \delta \mathbf{g}^{l}, \vec{\uptheta}^{l} \rangle  \nonumber \\
 \text{subject to }  \quad \vec{y}_{k+1}^l &= \vec{y}_{k}^l + \Delta_t^l F(\vec{y}_{k}^l, \vec{\uptheta}_k^l), \label{eq:level_dependent_problem}  \\
				\vec{y}_{0}^l &= \mat{Q} \vec{x}, \nonumber
\end{align}
where  $\delta \vec{g}^{l}$ is given by
\begin{align*}
\delta \vec{g}^{l}:=
\mat{R}^l  \nabla \mathcal{H}^{l+1}(\vec{\uptheta}^{l+1, \mu^l_1}, \vec{y}^{l+1, \mu^l_1}_K) - \nabla \mathcal{L}^l(\vec{\uptheta}^{l,0}, \vec{y}^{l, 0}_K),   
\label{eq:delta_g}
\end{align*}
for all levels $l < L$. 
For the finest level, $l=L$, we assume that $\delta \vec{g}^{L}:= \vec{0}$, and therefore  the functional $\mathcal{H}^l: \R^{n^l} \rightarrow \R$ coincides with the loss functional $\mathcal{L}^L$ defined in \eqref{eq:discrete_problem}. 

On the coarser levels, the functional $ \mathcal{H}^l $ consists of two terms: the loss functional $\mathcal{L}^l$ and the so-called coupling term $ \langle \delta \vec{g}^{l}, \vec{\uptheta}^{l} \rangle$. 
The coupling term creates a connection between two subsequent levels of the multilevel hierarchy. 
This is accomplished using the $\delta \vec{g}^l$ term, which measures the deviation between the restricted fine-level gradient $\nabla \mathcal{H}^{l+1}(\vec{\uptheta}^{l+1, \mu^l_1},  \vec{y}^{l+1, \mu^l_1}_K)$, and the initial coarse-level gradient $\nabla \mathcal{L}^l(\vec{\uptheta}^{l,0}, \vec{y}^{l, 0}_K)$.
The use of this coupling term is of major importance, as it enforces the following relationship:
\begin{align*}
\nabla \mathcal{H}^{l}(\vec{\uptheta}^{l,0}, \vec{y}^{l, 0}_K)  = \mat{R}^l  \nabla \mathcal{H}^{l+1}(\vec{\uptheta}^{l+1, \mu^l_1}, \vec{y}^{l+1, \mu^l_1}_K)
\end{align*}
for the first optimizer step on a given level. 
In addition, it guarantees, that the minimization on the coarse level is guided by the restricted fine level gradient and that the prolongated coarse level correction will be a descent direction on the fine level \cite{Nash2000multigrid}.

\subsubsection{Multilevel gradient-based methods}
\label{sec:ml_gd_methods}
Algorithm \ref{alg:mgopt} employs an auxiliary optimizer on every level of the multilevel hierarchy. 
By design, we are conveniently independent in the choice of the optimizer on each level. 
Our multilevel framework does not even require to employ the same type of optimizer on all levels. 
One can, for example, utilize computationally expensive optimizers on the coarser levels, while employing computationally cheaper optimizers on the finer levels.
In the multilevel community, it is quite popular to employ a second-order optimizer on the coarsest level and gradient-based optimizers on all finer levels. 

The easiest way to construct a multilevel training algorithm is to employ a gradient method on all levels. 
One level-optimizer iteration then consists of a simple gradient step computed using the whole dataset, see Algorithm \ref{alg:ml_alg_gd}.
Although, the choice of other gradient-based algorithms, such as LBFGS \cite{le2011optimization}, or RMSprop \cite{tieleman2012lecture}, might be more beneficial, using a vanilla gradient descent method allows for plain testing of our multilevel framework without introducing additional hyper-parameters. 
\begin{algorithm}[H]
	\caption{V-cycle of MG/OPT($\mathcal{L}^l, l, \vec{\uptheta}^{l,0}, \delta \mathbf{g}^{l} $) }
	\label{alg:mgopt}
	\begin{algorithmic}
	\CONSTANTS $  \mu_1^l, \mu_2^l, \mu^1 \in \N$
		\STATE \textit{1. Downward phase}
		\STATE \quad Construct  $\mathcal{H}^{l}$ by means of  \eqref{eq:level_dependent_problem} 
		\STATE \quad  $[\vec{\uptheta}^{l,\mu_1^l}]$ =  LevelOptimizer($\mathcal{H}^{l}$, $\vec{\uptheta}^{l,0}$, $\mu_1^l$)
		\STATE \quad $\vec{\uptheta}^{l-1, 0} \mapsfrom \mat{R}^{l-1} \vec{\uptheta}^{l, \mu_1^l}$ 	
		\STATE \quad Evaluate $\delta \mathbf{g}^{l-1}$
		\STATE \textit{2. Recursion or call to optimizer on the coarsest level}
		 \STATE \quad \textbf{if} $l = 2$ \textbf{then}
		 \STATE \quad \quad $[\vec{\uptheta}^{l-1,\mu^l}]$ =  LevelOptimizer($\mathcal{H}^{1}$, $\vec{\uptheta}^{l-1,0}$, $\mu^{1}$)
		 \STATE \quad \textbf{else}
		 \STATE  \quad \quad $[\vec{\uptheta}^{l-1,\mu^l}]$ = MG/OPT($l-1, \vec{\uptheta}^{l-1,0}, \delta \mathbf{g}^{l-1} $) 
		 \STATE \quad \textbf{end if}
		\STATE \textit{3. Upward phase}		 
		\STATE \quad $\vec{e}^{l-1} \mapsfrom \vec{\uptheta}^{l-1,\mu^l} - \vec{\uptheta}^{l-1,0}$
		\STATE \quad $\vec{e}^{l} \mapsfrom \I^{l-1} \vec{e}^{l-1}$ 
		\STATE \quad $\vec{\uptheta}^{l, \mu^l_1 +1} \mapsfrom \vec{\uptheta}^{l, \mu_1^l} + \vec{e}^{l}$ 
		\STATE \quad $[\vec{\uptheta}^{l,\mu^l}]$ =  LevelOptimizer($\mathcal{H}^{l}$, $\vec{\uptheta}^{l,\mu^l +1}$, $\mu_2^l$)
		\RETURN $\vec{\uptheta}^{l, \mu^l}$
	\end{algorithmic}
\end{algorithm}

\begin{algorithm}
	\caption{LevelOptimizer($\mathcal{H}^l$, $\vec{\uptheta}^{l,0}$, max\_it)}
	\label{alg:ml_alg_gd}
	\begin{algorithmic}[1]
		\CONSTANTS  $\alpha\in \R^+$	
		\FOR{ $i = 1, \dots, \text{max\_it}$}
		\STATE $\vec{\uptheta}^{l,i} = \vec{\uptheta}^{l,i-1} - \alpha \nabla \mathcal{H}^l(\vec{\uptheta}^{l,i-1}, \vec{y}^{l, i-1}_K)$
		\ENDFOR
		\RETURN $\vec{\uptheta}^{l, \text{max\_it}}$
	\end{algorithmic}
\end{algorithm}

\paragraph{Multilevel mini-batch gradient descent}
Since mini-batch gradient descent (SGD) is typically the algorithm of choice when training a neural network, here we propose its multilevel variant, Algorithm \ref{alg:sgd_loop}. 
Similarly to the single level SGD algorithm, we split the dataset into $nb$ mini-batches.
The algorithm then iterates through all mini-batches.
For each mini-batch, the algorithm invokes a multilevel gradient descent step, thus a V-cycle of MG/OPT configured with a gradient descent optimizer on all levels. 
\begin{algorithm}[H]
	\caption{One epoch of multilevel SGD}
	\label{alg:sgd_loop}
	\begin{algorithmic}[1]
		\CONSTANTS  $nb, L \in \N$	
		\FOR{ $b = 1, \dots, nb$}
		\STATE Construct $\mathcal{L}^L$ using mini-batch $\mathcal{D}_b$
		\STATE $[\vec{\uptheta}^{L, b}]$ = MG/OPT($\mathcal{L}^L, L, \vec{\uptheta}^{L, b-1}, \vec{0}$) 
		\ENDFOR
		\RETURN $\vec{\uptheta}^{L, nb}$
	\end{algorithmic}
\end{algorithm}

\paragraph{Computational complexity}
One V-cycle of the multilevel training strategy is computationally more expensive than one iteration of a single level optimizer. 
For the gradient-based optimizers, the computational cost is associated with the evaluation of the gradient, thus with the cost of a forward-backward pass.
To provide a fair comparison between multilevel and single level methods, we introduce the notation of work units. 
One work unit $\mathcal{U}^L$ represents the cost of a gradient evaluation on the finest level. 
Assuming a coarsening factor of two, the cost related to the gradient evaluation on the coarser levels is $\mathcal{U}^l=2^{l-L}\mathcal{U}^L$. 
The computational cost of one V-cycle, denoted $\mathcal{U}_c$, can be obtained by summing over the cost required on each level, thus
\begin{equation}
\begin{aligned}
\mathcal{U}_c \approx  \big(2^{1-L} \mu^0 +  \sum_{l=2}^{L}  (\mu_1^l + \mu_2^l + 1)  2^{l-L} \big)\mathcal{U}^L . 
\end{aligned}
\end{equation}
The cost on the coarsest level is related to $\mu^0$ level-optimizer steps.
On all other levels, we have to take into account the gradient evaluation that is required for computing the coupling term $\delta \vec{g}^l$, in addition to 
$\mu_1^l$ and $\mu_2^l$ level-optimizer steps.
The overall computational cost of the multilevel training $\mathcal{U}$ is then simply computed as
$\mathcal{U} = (\#\text{V-cycles}) \ \mathcal{U}_c$, where $(\#\text{V-cycles})$ denotes the number of V-cycles required to achieve a prescribed tolerance.

 \section{Numerical Experiments}
\label{sec:num_results}
We analyze the performance of the proposed multilevel optimizers using two classification problems:
\begin{itemize}[noitemsep, topsep=0pt]
\item \textbf{Co-centric circles}:
This simple example was proposed in \cite{lin2018resnet} and requires the classification of particles into two distinct classes. 
The input features $\vec{x}_j \in [-3, 3]^2$ describe the position of a particle in a two-dimensional plane, while the output vector $\vec{c}_j \in \R^2$ prescribes an affiliation to a given class. 
In particular, the $i$-th element of label $\vec{c}_j$ is defined as follows:
\begin{equation}
(\vec{c}_j)_i  =
\begin{cases}
1 & \text{if}  \quad 2 \leq  \| \vec{x}_j  \| < 3,\\
0 & \text{otherwise}. 
\end{cases}
\label{co-centric_circles_eq}
\end{equation}
The dataset consists of $3,000$ samples,  where $2,000$ are used for training and $1,000$ for testing.
\item \textbf{MNIST}:
Our second classification task considers the database of handwritten digits \cite{lecun1998gradient}. 
The dataset contains greyscale images of size $28 \times 28$ pixels that are uniformly divided into ten classes.
As a preprocessing, we standardize the images, so pixel values lie in the range $[0,1]$, and perform centering by subtracting the mean from each pixel. 
The data is split into $60,000$ samples for training and $10,000$ samples for testing.
\end{itemize}

\begin{figure}[htb]
\centering
	\begin{tikzpicture}
	\pgfplotsset{
		width=0.6\columnwidth,
		scale only axis,
		separate axis lines,
		every x tick label/.append style={font=\footnotesize},
		legend style={at={(1,0.76)},legend cell align=left,align=left,fill=none, draw = none, font=\footnotesize},
	}
	
	\begin{axis}[
	axis y line*=left,
	ymode=normal, 
	xmin = 1, xmax = 500,
	xmode=log,
	ymin=0.7,
	ymax=1.5,
	xlabel={\# V-cycles},
	ylabel=Training loss,
	y label style={at={(-0.14,0.5)}},
	legend entries={{1 level},{2 levels}, {4 levels}, {8 levels}}, 
	font=\footnotesize
	]
	
	\addplot[color = violet, very thick] 
	table [x={V-cycle}, y={total train loss}, col sep=comma] {circular_1lev.csv};
	\addplot[color = myyellow, very thick] 
	table [x={V-cycle}, y={total train loss}, col sep=comma] {circular_2lev.csv};
	\addplot[color = green2, very thick] 
	table [x={V-cycle}, y={total train loss}, col sep=comma] {circular_4lev.csv};
	\addplot[color = myred, very thick] table [x= {V-cycle}, y={total train loss}, col sep=comma] {circular_8lev.csv};
	\end{axis}
	
	\begin{axis}[
	axis x line=none,
	xmin = 1,
	xmax = 500,
	ymin=0.4,
	xmode = log,
	axis y line*=right,
	set layers,axis background,
	ylabel=Validation accuracy,			
	every y tick label/.append,
	style={font=\footnotesize}
	]
	
	\addplot[color = violet, dashed, very thick] 
	table [x={V-cycle}, y={validation accuracy}, col sep=comma] {circular_1lev.csv};
	\addplot[color = myyellow, dashed, very thick] table [x=V-cycle, y=validation accuracy, col sep=comma]  {circular_2lev.csv};
	\addplot[color = green2, dashed, very thick] table [x=V-cycle, y=validation accuracy, col sep=comma] {circular_4lev.csv};
	\addplot[color = myred, dashed, very thick] table [x= V-cycle, y=validation accuracy, col sep=comma]  {circular_8lev.csv};
	\end{axis}
	\end{tikzpicture}
	\begin{tikzpicture}
	\pgfplotsset{
		width=0.6\columnwidth,
		scale only axis,
		separate axis lines,
		every x tick label/.append style={font=\footnotesize},
		legend style={at={(0.95,0.75)},legend cell align=left,align=left,fill=none, draw = none, font=\footnotesize},
	}
	
	\begin{axis}[
	axis y line*=left,
	xmin = 1,
	ymin=0,
	ymax=3.2,
	xmode = log,
	ymode=normal,
	xlabel={\# V-cycles},
	ylabel=Training loss,
	legend entries={{1 level},{2 levels}, {4 levels}, {8 levels}}, 
	font=\footnotesize
	]
	\addplot[color = violet, very thick] table [x=V-cycle, y=total train loss, col sep=comma]
	{mnist_1lev.csv};		
	\addplot[color = myyellow, very thick] table [x= V-cycle, y=total train loss, col sep=comma] 
	{mnist_2lev.csv};		
	\addplot[color = green2, very thick] table [x= V-cycle, y=total train loss, col sep=comma] 
	{mnist_4lev.csv};					
	\addplot[color = myred, very thick] table [x= V-cycle, y=total train loss, col sep=comma]{mnist_8lev.csv};		
	\end{axis}
	
	\begin{axis}[
	axis x line=none,
	xmin = 1,
	ymax=1.1, 
	xmode = log,
	ymode=normal,
	axis y line*=right,
	ylabel=Validation accuracy,			
	every y tick label/.append,
	style={font=\footnotesize}
	]
	
	\addplot[color = violet, very thick, dashed] table [x=V-cycle, y=validation accuracy, col sep=comma]
	{mnist_1lev.csv};
	\addplot[color = myyellow, dashed, very thick] table [x= V-cycle, y=validation accuracy, col sep=comma] 
	{mnist_2lev.csv};
	\addplot[color = green2, dashed, very thick] table [x= V-cycle, y=validation accuracy, col sep=comma] 
	{mnist_4lev.csv};
	\addplot[color = myred, dashed, very thick] table [x= V-cycle, y=validation accuracy, col sep=comma]	{mnist_8lev.csv};
	\end{axis}
	\end{tikzpicture}
\caption{
The loss function (solid lines) and validation accuracy (dashed lines) as a function of V-cycles.
Results obtained for varying numbers of levels.
Top: Co-centric circles. Bottom: MNIST.
}
\label{fig:wrt_levels}
\end{figure}
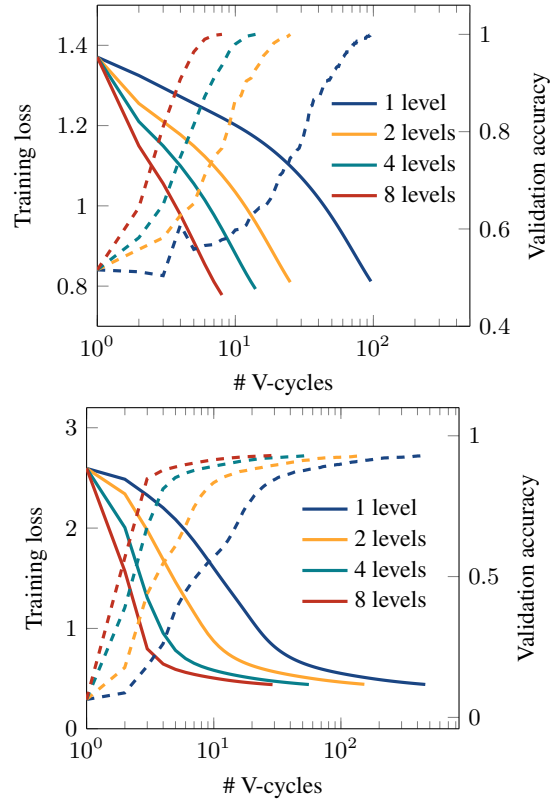 
\paragraph{Implementation and testing environment}
Our implementation of deep residual networks and  nonlinear multilevel training framework uses the Keras \cite{chollet2015keras} and Tensorflow \cite{tensorflow2015-whitepaper} library.

The classification is performed using a ResNet architecture as described in \eqref{eq:res_layer}.
We employ a simple variant of residual blocks, i.e. a one layer perceptron with ReLu activation function, see \eqref{eq:f_tilde}.
Each layer consists of $3$ nodes for the co-centric circles example and $10$ nodes for the MNIST example. 
The operator $\mat{Q}$, which maps input features into network width, is learned during training. 
All layers are fully-connected.

Unless specified differently, the deep residual network consists of $2,048$ residual blocks. 
In the case of multilevel training, the multilevel-hierarchy of auxiliary networks with different resolutions is created by coarsening in time with a factor of two, see Section \ref{sec:discretization} for more details.  
On each level of the multilevel hierarchy, we consider the final time $T=1$. 
As commonly used, we employ Tikhonov regularization, thus $\mathcal{R}(\cdot): = 
\beta \| \cdot \|_F^2$, where $ \| \cdot \|_F^2$ denotes the Frobenius norm.
On the finest level (deepest network), we prescribe the regularization parameters $\beta=10^{-4}$ and  $\beta=10^{-5}$ for co-centric circles and MNIST, respectively. 
On the coarser levels, the value of the regularization parameter $\beta$ is scaled by a coarsening factor $2^{L-l}$, where $l$ denotes a given level. 

All presented experiments were performed on our local cluster consisting of $42$ compute nodes, each equipped with 2 Intel R E5-2650 v3 processor with a clock frequency of 2.60 GHz.  The memory per node is 64 GB.

\paragraph{Algorithmic setup}
We trained both test examples using gradient-based multilevel optimizers with a constant learning rate of $0.1$ in the co-centric circle example and $0.01$ in case of the MNIST dataset.
During all numerical tests, the weights are initialized randomly, while biases are set to zero. 
The co-centric circles example is trained using the full dataset, giving rise to a multilevel gradient descent.
We terminate training if a validation accuracy of $1$ is achieved. 
The training of the MNIST example is performed using a multilevel mini-batch gradient descent with a mini-batch size of $1,000$.
As a termination criterion, we require the validation accuracy to be higher than $0.93$.

During multilevel training, level-optimizers perform several steps while completing the downward and upward phase of the V-cycle. 
To describe our particular level-optimizer setup, we introduce a list notation, such as $[(1), 2, 1, 3, \{4\}]$ for a $5$-level training strategy. 
Each entry of the list indicates a number of optimizer steps used on a given level.
The list is ordered from the finest to the coarsest level.
If no bracket is used, we assume $\mu_1^l=\mu_2^l$. 
The use of a regular bracket implies that the level-optimizer was not called during the upward phase, thus $\mu_2^l=0$. 
The curly bracket indicates the number of optimizer steps on the coarsest grid, i.e. $\mu^1$.

\begin{figure}[h]
\centering
	\begin{tikzpicture} 
	\pgfplotsset{
		width=0.6\columnwidth,
		scale only axis,
		separate axis lines,
		every x tick label/.append style={font=\footnotesize},
		legend style={at={(0.99,0.63)},legend cell align=left,align=left,fill=none, draw = none, font=\footnotesize},
	}
	
	\begin{axis}[
	axis y line*=left,
	ymode=normal, 
	xmode=log,
	xmin = 1,
	xmax = 100,
	xlabel={\# V-cycles},
	ylabel=Training loss,
	legend entries={64 res blks, 256 res blks, 2048 res blks}, 
	font=\footnotesize
	]
	\addplot[color = green2, very thick] table [x=V-cycle, y=total train loss, col sep=comma] {circular_64blks.csv};
	\addplot[color = myred, very thick] table [x=V-cycle, y=total train loss, col sep=comma] {circular_256blks.csv};
	\addplot[color = myyellow, very thick] table [x=V-cycle, y=total train loss, col sep=comma]{circular_2048blks.csv};
	\end{axis}
	\begin{axis}[
	axis x line=none,
	xmin = 1,
	xmax = 100,
	xmode = log,
	axis y line*=right,
	ylabel=Validation Accuracy,			
	every y tick label/.append,
	every x tick label/.append,	
	style={font=\footnotesize}
	]
	\addplot[color = green2, dashed, very thick] table [x=V-cycle, y=validation accuracy, col sep=comma] {circular_64blks.csv};
	\addplot[color = myred, dashed, very thick] table [x=V-cycle, y=validation accuracy, col sep=comma] {circular_256blks.csv};
	\addplot[color = myyellow, dashed, very thick] table [x=V-cycle, y=validation accuracy, col sep=comma] {circular_2048blks.csv}; 
	\end{axis}
	\end{tikzpicture}
\centering
	\begin{tikzpicture}
	\pgfplotsset{
		width=0.6\columnwidth,
		scale only axis,
		separate axis lines,
		every x tick label/.append style={font=\footnotesize},
		legend style={at={(0.95,0.65)},legend cell align=left,align=left,fill=none,
			draw = none, font=\footnotesize},
	}	
	\begin{axis}
	[
	axis y line*=left,
	ylabel = Training loss,
	legend cell align={left}, 
	legend entries={128 res blks, 512 res blks, 2048 res blks}, 
	xmin=1,
	xmode = log,
	ymode=normal,
	xmax = 100,
	ymin=0,
	ymax=3.2,
	xlabel={\# V-cycles},
	font=\footnotesize,
	]
	\addplot[color = green2,very thick] table [x=V-cycle, y=total train loss, col sep=comma] {mnist_128blks.csv};
	\addplot[color = myred, very thick] table [x=V-cycle, y=total train loss, col sep=comma]{mnist_512blks.csv};
	\addplot[color = myyellow, very thick] table [x=V-cycle, y=total train loss, col sep=comma] {mnist_2048blks.csv};
	\end{axis}
	
	\begin{axis}[
	axis x line=none,
	xmin = 1,
	ymax=1.1,
	xmode = log,
	ymode=normal,
	xmax=100,
	axis y line*=right,
	ylabel=Validation accuracy,			
	every y tick label/.append,
	every x tick label/.append,	
	style={font=\footnotesize}
	]
	\addplot[color = green2, dashed, very thick] table [x=V-cycle, y=validation accuracy, col sep=comma]{mnist_128blks.csv};
	\addplot[color = myred, dashed, very thick] table [x=V-cycle, y=validation accuracy, col sep=comma] {mnist_512blks.csv};
	\addplot[color = myyellow, dashed, very thick] table [x=V-cycle, y=validation accuracy, col sep=comma] {mnist_2048blks.csv};
	\end{axis}
	\end{tikzpicture}
\caption{
Loss function (solid lines) and validation accuracy (dashed lines) over number of V-cycles.
Results obtained for varying numbers of residual blocks (res blks).
Top: 6-level method trained on co-centric circles. 
Bottom: 8-level method trained on MNIST.
}
\label{fig:num_blocks}
\end{figure}
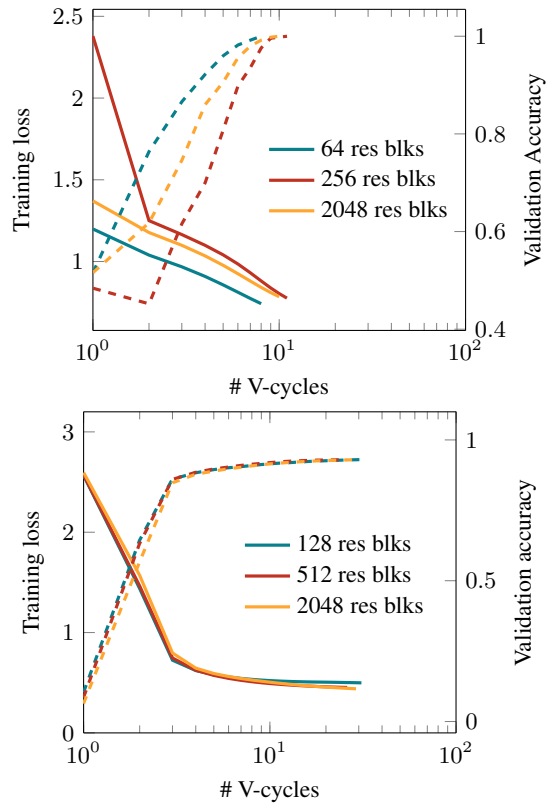 
\subsection{Numerical results}
\label{sec:wrt_levels}
\paragraph{Convergence behavior with respect to levels}
We analyze the convergence behavior of the proposed multilevel training methods with respect to varying numbers of levels. 
The presented results also include the single-level version of the algorithms, thus vanilla gradient descent and mini-batch gradient descent. 
During the experiments, we employ one iteration of level-optimizer for all levels $l \in \{L/2+1, \dots, L\}$. 
On the coarser levels, that is $l \in \{1, \dots, L/2\}$, we perform $2$ level-optimizer iterations. 
The described setup is used for both the downward and the upward phase of a V-cycle, except on the finest level, where we skip the call to the optimizer during the upward phase. 
Thus, for 6-level training, we use following setup $[(1),1,1,2,2,\{2\}]$.

\begin{table}[H]
	\centering
	\small
	\caption{Computational cost of multilevel training with respect to varying numbers of levels. 
	The symbol $ { \mathcal{U}_c}$ denotes the cost per V-cycle and  $\mathcal{U}$ stands for the total cost. }
	\label{fig:comp_cost}	
	\begin{tabular}{|c|l|c|c|c|c|}
		\multicolumn{5}{c}{Co-centric circle} \\ \hline
		 { L} & { Optimizers setup} & { \# V-cycles} & $ { \mathcal{U}_c}$ & { $\mathcal{U}$} \\ \hline
		1 &[\{1\}]	& 95 & 1.00 & 95 \\ \hline
		2 &[(1),\{2\}]& 32 & 3.00 & 96 \\ \hline
		4 &	[(1),1,2,\{2\}] & 12 & 5.00 & 60 \\ \hline
		6 & [(1),1,1,2,2,\{2\}]& 7 & 5.25 & 37 \\ \hline
		8 &	[(1),1,1,1,2,2,2,\{2\}]& 5 & 5.19 & 26 \\ \hline
		\multicolumn{5}{c}{} \\  		
		\multicolumn{5}{c}{MNIST} \\  \hline
		{ L} & { Optimizers setup} & { \# V-cycles} & $ { \mathcal{U}_c}$ & { $\mathcal{U}$} \\ \hline
		1 &[\{1\}]	& 460 & 1.00 & 460 \\ \hline
		2 &[(1),\{2\}]& 152 & 3.00 & 456 \\ \hline
		4 &	[(1),1,2,\{2\}] & 55 & 5.00 & 275 \\ \hline
		6 & [(1),1,1,2,2,\{2\}& 33 & 5.25 & 173 \\ \hline
		8 &	[(1),1,1,1,2,2,2,\{2\}]& 28 & 5.19 & 145 \\ \hline
	\end{tabular}
\end{table}

Figure \ref{fig:wrt_levels} demonstrates the obtained results for both test problems. 
As we can see, adding more levels reduces the number of required V-cycles significantly.
In particular, using the 2-level method already leads to a decrease in the number of iterations by a factor of $3$, compared to the single level method. 
For the 4-level method, we obtain a reduction in the number of V-cycles by a factor of 8, while for the 8-level method, the number of V-cycles is approximately reduced by a factor of $15$. 

We compare the total computational cost of the multilevel training methods.
Following the analysis presented in Section \ref{sec:ml_gd_methods}, we show the computational cost in Table \ref{fig:comp_cost} for both datasets. 
For more than four levels, the computational cost does not increase substantially anymore, as the cost of numerical operations on those levels is negligible compared to the cost of the same operations performed on the finest level.
The results also demonstrate, that the total computational cost $\mathcal{U}$ decreases as the number of levels increases. 
This is not surprising, as the number of V-cycles required for convergence decreased. 
In particular, the 8-level method is approximately $3.4$ times computationally more efficient than its single level counterpart.

\paragraph{Convergence behavior with respect to number of residual blocks}
Further, we analyze the convergence behavior of the proposed multilevel training strategy for varying numbers of residual blocks. 
We keep the same parameters as described in the beginning of Section \ref{sec:num_results}, but alter the number of residual blocks.
Figure \ref{fig:num_blocks} illustrates the obtained results for both datasets. 
As we can see, the method exhibits the same asymptotic convergence behavior independently on the number of residual blocks. 
These results are very promising, as they suggest that the convergence rate of our multilevel training strategy does not deteriorate with network depth. 

\paragraph{Influence of hyper-parameters}
In the end, we investigate the sensitivity of multilevel methods with respect to the choice of hyper-parameters. 
Firstly, we demonstrate how different setups of level-optimizers influence the convergence properties of the multilevel training strategy. 
Table \ref{fig:diff_iter_comb} reports the results obtained for different setups of the 8-level method trained on the co-centric circle example. 
As contemplated, increasing the number of optimizer calls on the lower levels decreases the total computational cost of the multilevel method. 
This is due to the fact, that additional calls to coarse level optimizers lower the number of required V-cycles.
But at the same time, those additional calls do not considerably increase the cost of one V-cycle.
The most expensive part of the V-cycle is the optimizer call on the finest level.
Therefore, it is beneficial to skip it during the upward phase. 
This has no substantial impact on the performance of the method, as the upward optimizer step is immediately followed by the downward optimizer step of the next V-cycle.

Secondly, we study the sensitivity with respect to the choice of learning rate and regularization parameter.
Here, we consider the co-centric circle example and three different values of learning rate, $\alpha=\{ 0.05; 0.1; 0.5\}$, and  regularization parameter $\beta=\{10^{-3}; 10^{-4}; 10^{-5} \}$.
This yields nine different hyper-parameter setups, which we tested using single level and 8-level ([(1),1,1,1,2,2,2,\{2\}]) methods. 
On average, the computational cost of the 8-level method is $3.54$ lower than the cost required by a single level method. 
The relative standard deviation of our results is $8.02 \%$.

\begin{table}[H]
	\centering
	\caption{
	Computational cost as a function of optimizer steps for co-centric circles example with the 8-level method.
	The symbol $ { \mathcal{U}_c}$ denotes the cost per V-cycle and  $\mathcal{U}$ stands for the total cost. 
		}	
	\label{fig:diff_iter_comb}
	{	\small
	\begin{tabular}{|l|c|c|c|c|}
		\hline 
		Optimizers setup & \# V-cycles & $\mathcal{U}_c$ & $\mathcal{U}$ \\ \hline 
		$[1,1,1,1,1,1,1,\{1\}]$	& 7 & 5.97 & 42\\ \hline
		$[1,1,1,1,1,1,1,\{2\}]$	& 7 &5.96 & 42 \\ \hline
		$[1,1,1,1,1,1,1,\{5\}]$	& 6 & 5.99 & 36 \\ \hline
		$[1,1,1,1,1,1,1,\{10\}]$ & 5 & 6.03 & 30 \\ \hline
		$[1,1,1,1,2,2,2,\{2\}]$	& 5 & 6.19 &31 \\ \hline
		$[(1),1,1,1,2,2,2,\{2\}]$& 5 & 5.19 & 26 \\ \hline
	\end{tabular}
	}
\end{table}
 \section{Conclusion}
\label{sec:conclusion}
In this work, we proposed a nonlinear multilevel minimization framework for training the deep residual networks.
Our multilevel framework is based on MG/OPT framework \cite{Nash2000multigrid} and utilizes a hierarchy of auxiliary networks with different depths to speed up the training process of the original network. 
Using our novel training framework, we proposed multilevel gradient and mini-batch gradient methods.
The performed numerical experiments demonstrated the convergence behavior of multilevel training methods and showed significant decrease in the computational cost compared to its single level variants.

\bibliographystyle{icml2020}

\end{document}